Using large language models to estimate features of multi-word expressions:

Concreteness, valence, arousal

Gonzalo Martínez [1]  Juan Diego Molero [2]  Sandra González [2]  Javier Conde [2]

Marc Brysbaert [3]  Pedro Reviriego [2]

[1] Universidad Carlos III de Madrid, Spain

[2] ETSI de Telecomunicación, Universidad Politécnica de Madrid, Spain

[3] Department of Experimental Psychology, Ghent University, Belgium

Correspondence address:     Marc Brysbaert
                            Department of Experimental Psychology
                            Ghent University
                            9000 Ghent, Belgium
                            marc.brysbaert@ugent.be

Abstract

This study investigates the potential of large language models (LLMs) to provide accurate estimates of concreteness, valence and arousal for multi-word expressions. Unlike previous artificial intelligence (AI) methods, LLMs can capture the nuanced meanings of multi-word expressions. We systematically evaluated ChatGPT-4o's ability to predict concreteness, valence and arousal. In Study 1, ChatGPT-4o showed strong correlations with human concreteness ratings (r = .8) for multi-word expressions. In Study 2, these findings were repeated for valence and arousal ratings of individual words, matching or outperforming previous AI models. Study 3 extended the prevalence and arousal analysis to multi-word expressions and showed promising results despite the lack of large-scale human benchmarks. These findings highlight the potential of LLMs for generating valuable psycholinguistic data related to multi-word expressions. To help researchers with stimulus selection, we provide datasets with AI norms of concreteness, valence and arousal for 126,397 English single words and 63,680 multi-word expressions.

Some words/expressions are easier to understand or produce than others. Language researchers examine the variables that affect processing ease, to build theories about the processes involved in understanding and producing language. A number of variables can be measured objectively, such as word frequency, part of speech or word length. Other variables are more subjective and are usually measured by asking people to give ratings. These variables include impressions of concreteness, age of acquisition, familiarity, or feelings associated with words/expressions.

Online testing has greatly facilitated the collection of ratings, so that it is now possible to collect ratings for thousands of words/expressions within weeks and at an affordable price (if one has access to grant money). As a result, many large-scale reviews have been published in recent years (e.g., Brysbaert et al., 2014; Diez-Alamo et al., 2019; Hinojosa et al., 2023; Proos & Aigro, 2023; Warriner et al., 2013).

At the same time, it has become clear that good estimates of ratings can be obtained with artificial intelligence (AI). At first, it did not work well because researchers tried to approximate ratings by taking average values of closely related words (Mandera et al., 2015), but soon researchers discovered that better results were obtained when they worked with the semantic vectors of individual words. Hollis et al. (2017; see also Westbury et al., 2015) used the Google Word2vec semantic vectors (Mikolov et al., 2013) as predictors of human ratings of concreteness, valence and arousal in a linear regression and showed that the predictions correlated 0.8 with the ratings obtained. More importantly, the predictions generalized to items not used in the initial model estimation (cross-validation), so the model could be used to estimate values for all 78,286 words with semantic vectors in the English word list they used.

The work of Hollis et al. (2017) was extended to large language models based on deep learning with multiple hidden layers between input and output (Plisiecki & Sobieszek, 2023; Solovyev et al., 2022; Wang & Xu, 2023). The approach has also been used successfully to estimate values for under-resourced languages through translation (Buechel et al., 2020; Thompson & Lupyan, 2018).

Almost all available research is limited to single words. This is a restriction as half of language utterances consist of multiword expressions (Biber et al., 2004; Conklin & Schmitt, 2008; Muraki et al., 2023). These include compound nouns (bird watcher, blind luck), particle verbs (throw up, zone in), and fixed expressions (a drop in the ocean, good morning). There is evidence that multiword expressions are not simply understood by parsing the words, but are stored as separate entities in the mental lexicon, just like single words. This can be concluded from the finding that multiword expressions are processed faster than other matched word sequences, and are influenced by factors such as the frequency and age of acquisition of the multiword expression (Arnon & Snider, 2020; Senaldi et al., 2022; Yi & Zhong, 2024).

Muraki et al. (2023) put multiword expressions on the map of psycholinguistic norms by collecting concreteness ratings for 62,000 English multiword expressions. A logical next step would be to collect ratings for more variables, but given the evidence that large language models predict human ratings of individual words, it is worth exploring how well they estimate multi-word expressions. A major limitation up to the introduction of large language models was that information was limited to single words (e.g., semantic vectors). Since large language models no longer work with single words as input and output, they can potentially provide useful information for word sequences, even if these sequences can take many different forms (such as wash yourself, washing oneself, washes himself, washed themselves, ...).

The present study has two goals. First, we want to see how well ratings from large language models can predict human ratings of multiword expressions as a proof of concept. We do this by comparing how well ChatGPT-4o estimates of concreteness approximate the human ratings collected by Muraki et al. (2023). Second, we aim to provide estimates of valence and arousal to researchers. Sentiment analysis and research into the processes involved in emotional language processing occupy a prominent place in current research (de Zubicaray & Hinojosa, 2024; Wankhade et al., 2022), and the possibility of extending this research to multiword expressions will help make progress. The first goal is addressed in Study 1; the second goal in Study 3.

Study 1: Predicting the Muraki et al. (2023) ratings with ChatGPT-4o

We conducted pilot tests on small samples of expressions with several large language models (including models freely available for research), but we discuss only the results of ChatGPT-4o (Open AI, 2023) because they were better than the other models we tried. We used the latest versions[1] that was available via the Application Programming Interface (API) in July-August 2024.

We also tried several types of instructions. At first, we used the instructions provided by Muraki et al. (2023), but found they were too verbose (363 words) to be repeated each time a rating for an expression was asked from the model. We also tried a version in which the model was provided with two expressions and asked to indicate which one was more concrete. In the end, we found the best results with the instructions in prompt 1 below (the three examples of stimuli at each end of the scale were taken from Brysbaert et al. (2014) and slightly improve the correlations with human ratings):

Prompt1

"Could you please rate the concreteness of the following multiword expression on a scale from 1 to 5, where 1 means very abstract and 5 means very concrete? Examples of words that would get a rating of 1 are essentialness, although and hope. Examples of words that would get a rating of 5 are bat, frangipane, and blackbird. The expression is: [insert expression here]. Only answer a number from 1 to 5. Please limit your answer to numbers."

We set the temperature to zero, so that the same results would be obtained in replications (the temperature is a measure of the randomness added to the model to obtain a varied output on different trials; it is also summarized as the degree of creativity given to the model). The ChaptGPT-4 API returned two types of information. First, it gave a number from the 1-5 scale. Second, it returned an estimated probability of each response alternative[2]. For instance, the rating with the highest probability for "shoot a film" was 4 (prob = .646), followed by 3 (prob = .346), 5 (prob = .006), and 2 (prob = .001). This gave an estimated overall rating of 3.66.

---

[1] The version we used for the master lists of norms was "gpt-4o-2024-08-06". We noticed that the estimates of this version differed slightly from the gpt-4o-2024-05-13 version used in Studies 1-3, without changing the overall quality of the estimates.
[2] An introduction to LLMs and the output they provide is given in https://poloclub.github.io/transformer-explainer/.

Only the 62,889 expressions known by the participants of Muraki et al. (2023) were included, in order not to introduce noise due to obscure or uninterpretable entries[3]. The instructions were repeated for each expression to prevent response dilution.

|  | Muraki | GPT rating | GPT probs |
|---|---|---|---|
| Muraki |  | .798 | .812 |
| GPT rating | .793 |  | .988 |
| GPT probs | .807 | .978 |  |

**Table 1: Correlations between the human ratings of Muraki and the ChatGPT-4o estimates (rating with the highest probability and sum of the ratings times their probabilities). Above the diagonal: Pearson correlations; below the diagonal: Spearman correlations. (N = 62,889)**

Table 1 shows the correlations between the ChatGPT-4o estimates and the human ratings of Muraki et al. (2023). Given that the reliability of the Muraki et al. (2023) ratings is estimated at r = .84, the observed correlations of .8 comes close to the maximum value that can be expected. The sum of the ratings times the probabilities gave slightly more information than the rating with the highest probability (the same was true for all other analyses we ran). So, we will use this measure in all analyses below.

A more stringent test of whether the LLM estimates are equivalent to human ratings is by looking at the score distributions. As shown in Figure 1, they are not completely comparable. ChatGPT-4o gives more extreme values than humans and has modes around the integers of the Likert scale. Therefore, a better way to use LLM ratings as an alternative for human ratings may be to use rank scores rather than raw scores. Indeed, the Spearman correlation between the two variables is close to the Pearson correlation (compare the lower half to the upper half of Table 1).

---

[3] A risk of working with large databases is that some of stimuli are not known to participants or do not make sense to them. Examples of such stimuli are "1 timothy", "a cat can look at a king" or "à la provençale". Researchers interested in these specific expressions can use the instructions we used to get estimates from ChatGPT-4o (or other LLMs).

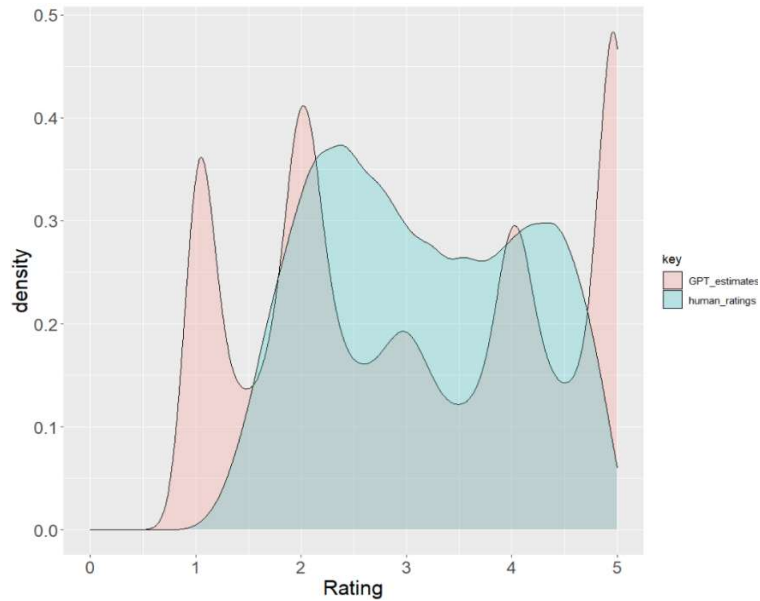

**Figure 1: Distribution of the human concreteness ratings and the ChatGPT-4o estimates for the 62,889 multiword expressions of Muraki et al. (2023).**

We would expect ChatGPT-4o to have the most difficulties with idioms in which the meaning of the multiword expression is much more abstract than the meanings of the words used, such as "a golden key can open any door". The human rating of this expression was 1 (Muraki et al., 2023); the ChatGPT-4o estimate was 2, which is closer than estimates obtained with previous AI tools.

To more systematically map the estimates of opaque idioms, we used the list of 525 most frequent idioms collected by Hsu (2020). There were 486 idioms with exactly the same wordings in our list of ChatGPT-4o estimates[4]. These gave a correlation of r = .56, considerably lower than the .81 observed in Table 1, partly due to range restriction (most estimates were low). Interestingly, of the expressions with a difference larger than 1.75 between the human ratings and the ChatGPT-4o estimates, there were nine in which the ChatGPT estimate was lower than the human ratings (serve notice, for my money, think twice, stay the course, hold water, fourth estate, above board, pull no punches, sit on the fence), whereas there was only one in which ChatGPT was higher (fat cat). So, it is not the case that ChatGPT-4o overestimates the concreteness of opaque idioms that use vivid metaphors to express more abstract ideas. All values for the Hsu idioms can be found on https://osf.io/k5a4x/.

Another advantage of ChatGPT-4o assessments is that they can be extended easily to new stimuli. Among the 791 new expressions collected since Muraki et al. (2023) were expressions such as "accede to", "high air pressure", and "coughing up blood." Estimates for these stimuli can be obtained without much effort from GPT (which yields values of 1.00, 3.12 and 4.85, respectively).

---

[4] As indicated in the introduction, an issue with multiple word expressions is that they can be summarized in different wordings. For instance, Hsu (2020) listed "a blind spot", whereas in our list we had "blind spot" (human rating = 2.7; GPT estimate = 3.17).

Study 2: Obtaining ChatGPT-4o word estimates for valence and arousal

Now that we know that useful estimates of word features of multiword expressions can be obtained with ChatGPT-4o, it becomes interesting to collect other norms. Two norms often used in research are valence and arousal. Valence refers to whether an expression communicates a positive or negative emotion. In the valence ratings of Warriner et al. (2013), words like pedophile and rapist got valence ratings of 1.3, whereas words like happiness and vacation got valence ratings of 8.5 (on a Likert scale of 1-9). Arousal refers to the degree of excitement that a stimulus evokes. Words with the lowest arousal ratings in Warriner et al. (2023) were grain, calm and dull; words with the highest ratings were sex, gun and insanity.

Information about valence and arousal is important to know to what extent information processing is influenced by the emotional nature of the stimulus materials (e.g., de Zubicaray & Hinojosa; Kyröläinen et al., 2021). It is also interesting for sentiment analysis of texts and messages (Birjali et al., 2021; Wankhade et al., 2022).

We used a two-step procedure to obtain valence and arousal estimates for our list of multiword expressions. First, we built on the 13,914 single-word ratings of Warriner et al. (2013) to decide which instructions were the best. As Warriner et al. used a rating scale from 1 to 9, we used the same scale. These were the final instructions we used for valence (Prompt 2):[5]

> Prompt 2
>
> "Could you please rate how reading the following multiword expression makes a person feel. Use a scale from 1 to 9, where 1 means very negative, bad and 9 means very positive, good. Examples of words that would get a rating of 1 are pedophile, AIDS and wreck. Examples of words that would get a rating of 9 are vacation, fantastic, and laugh. The expression is: [insert expression here]. Only answer a number from 1 to 9. Please limit your answer to numbers".

An advantage of valence and arousal estimates is that we have several sources to compare the output with. The ones we used were:

1. Warriner et al. (2013): human ratings on 1-9 Likert scale (N = 13,914).
2. Scott et al. (2019): human ratings on 1-9 Likert scale (N = 4,083 words in common with Warriner et al.).
3. Mohammad (2018): human ratings based on most–least (best–worst) rankings (N = 13,864 words in common with Warriner et al.).
4. Recchia & Louwers (2015): estimates based on similarity of ratings to ANEW words (N = 13,783 in common).
5. Hollis et al. (2017): estimates based on semantic vectors and Warriner et al. (2013) (N = 13,789 in common).
6. Buechel et al. (2020): estimates based on semantic vectors fine-tuned on Warriner et al. (2013) (N = 13,810 in common).

---
[5] We also tried the reverse order of ratings, going from 9 to 1. This gave slightly lower (.01-.02) correlations with the human ratings. The average of both instructions did not have higher validity either.

7. Plisiecki & Sobieszek (2024): estimates based on GPT-3 fine-tuned on Warriner et al. (2013) (N = 13,680 in common).

Table 2 shows the correlations we obtained. They are on par with the correlations obtained in other studies with large language models and better than the early estimates of Recchia & Louwers (2015) and Hollis et al. (2017). The Buechel and Plisiecki estimates did better on the Warriner et al. data on which they were fine-tuned, but not on the other two sets of human data.

|  | Warriner | Scott | Mohammad | Recchia | Hollis | Buechel | Plisiecki | GPT4 |
|---|---|---|---|---|---|---|---|---|
| Warriner |  | .92 | .86 | .80 | .84 | .93 | .97 | .90 |
| Scott | .91 |  | .90 | .80 | .84 | .91 | .91 | .93 |
| Mohammad | .84 | .88 |  | .77 | .81 | .87 | .86 | .87 |
| Recchia | .77 | .76 | .74 |  | .78 | .83 | .80 | .79 |
| Hollis | .81 | .82 | .79 | .74 |  | .89 | .84 | .83 |
| Buechel | .91 | .90 | .85 | .80 | .87 |  | .93 | .90 |
| Plisiecki | .87 | .89 | .83 | .78 | .81 | .91 |  | .89 |
| GPT4 | .88 | .92 | .86 | .76 | .81 | .89 | .88 |  |

**Table 2: Valence: Correlations between the human data (Warriner, Scott, Mohammad), estimates based on semantic vectors (Recchia, Hollis), and estimates based on LLM (Buechel, Plisiecki, GPT4). Upper half: Pearson correlations; lower half: Spearman correlations.**

The high correlation between AI estimates and human ratings hides the fact that the distributions of values differ considerably, as can be seen in Figure 2. Again, for some purposes it may be better to use ChatGPT-4o ranks rather than the raw values obtained. The lower half of Table 2 shows the Spearman correlations, which are only slightly lower than the Pearson correlations.

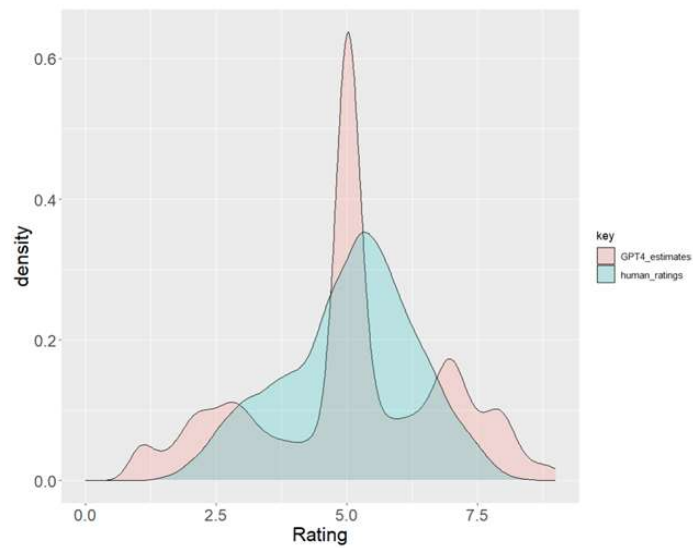

**Figure 2: Distribution of the human valence ratings and the ChatGPT-4o estimates for the 13,914 words tested in Warriner et al. (2013).**

Table 3 and Figure 3 show the same data for the arousal ratings. The prompt used was:

Prompt 3

"Could you please rate how reading the following multiword expression makes a person feel. Use a scale from 1 to 9, where 1 means very calm, relaxed and 9 means very aroused, energized. Examples of words that would get a rating of 1 are grain, dull and rest. Examples of words that would get a rating of 9 are gun, lover, and thrill. The expression is: [insert expression here]. Only answer a number from 1 to 9. Please limit your answer to numbers".

|          | Warriner | Scott | Mohammad | Recchia | Hollis | Buechel | Plisiecki | GPT4 |
|----------|----------|-------|----------|---------|--------|---------|-----------|------|
| Warriner |          | .61   | .68      | .62     | .68    | .79     | .92       | .74  |
| Scott    | .60      |       | .54      | .51     | .56    | .58     | .60       | .56  |
| Mohammad | .66      | .53   |          | .68     | .71    | .76     | .69       | .81  |
| Recchia  | .59      | .49   | .64      |         | .72    | .76     | .65       | .71  |
| Hollis   | .66      | .55   | .68      | .68     |        | .83     | .71       | .78  |
| Buechel  | .78      | .60   | .74      | .73     | .81    |         | .81       | .85  |
| Plisiecki| .92      | .59   | .66      | .60     | .68    | .80     |           | .76  |
| GPT4     | .73      | .57   | .79      | .68     | .77    | .83     | .74       |      |

Table 3: Arousal: Correlations between the human data (Warriner, Scott, Mohammad), estimates based on semantic vectors (Recchia, Hollis), and estimates based on LLM (Buechel, Plisiecki, GPT4). Upper half: Pearson correlations; lower half: Spearman correlations.

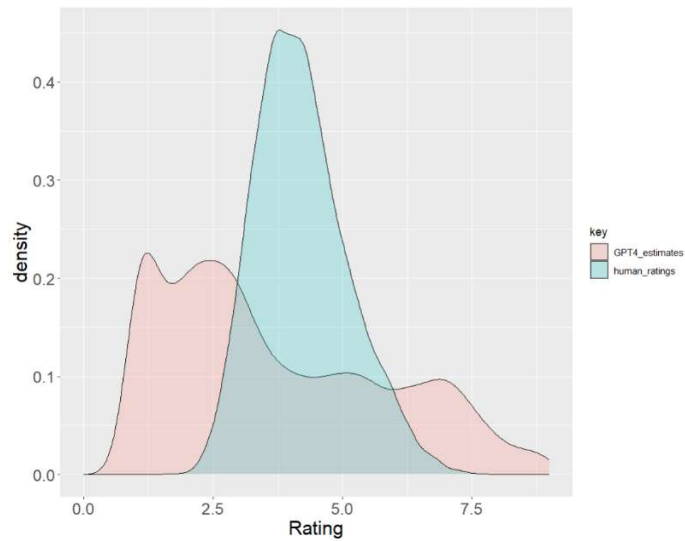

Figure 3: Distribution of the human arousal ratings and the ChatGPT-4o estimates for the 13,914 words tested in Warriner et al. (2013).

There is less agreement about the degree of arousal evoked by words, as can be seen in the lower correlations between the three sets of human data (Warriner, Scott, Mohammad). ChatGPT-4o does well relative to the other sources, in particular for the data of Scott and Mohammad, which were not used to fine-tune the models. The GPT estimates were more spread than the Warriner ratings and situated more at the low end of the arousal scale.

Study 3: Obtaining ChatGPT-4o multiword expression estimates for valence and arousal

Now that we have good instructions, we can obtain GPT estimates of valence and arousal for the multiword expressions of Muraki et al. The instructions for valence were the same as prompt 2 above; the instructions for arousal were those of prompt 3. To these, we added a list of 791 extra expressions we collected since.

The 10 expressions with lowest valence ratings were: child pornography, racial extermination, child molester, gang rape, paedophile ring, suicide bombing, ethnic cleansing, nazi party, child abuse, and white supremacy.

The 10 expressions with the highest valence ratings were: I love you, summer vacation, best friend, pure joy, Merry Christmas and a Happy New Year, on cloud nine, totally awesome, vacation time, Heaven on earth, perfect in every way.

The 10 expressions with the lowest arousal ratings were: of a, soybean oil, of an, rye seed, oat grass, such as, was to, oat bran, haricot bean, there are.

The 10 expressions with the highest arousal ratings were: gang rape, suicide bomber, racial extermination, suicide bombing, child pornography, child molester, terrorist act, terrorist attack, mass murder, suicide terrorist.

A further way to verify the quality of the estimates is to look at the relationship between valence and arousal. In human ratings there is an inverted U-shape relationship, because words with low and high valence have higher arousal ratings than words with medium valence ratings. Figure 4 shows that this is the case for the multiword expression estimates as well. In particular, negative words have a high arousal (see Warriner et al. for a similar pattern in people).

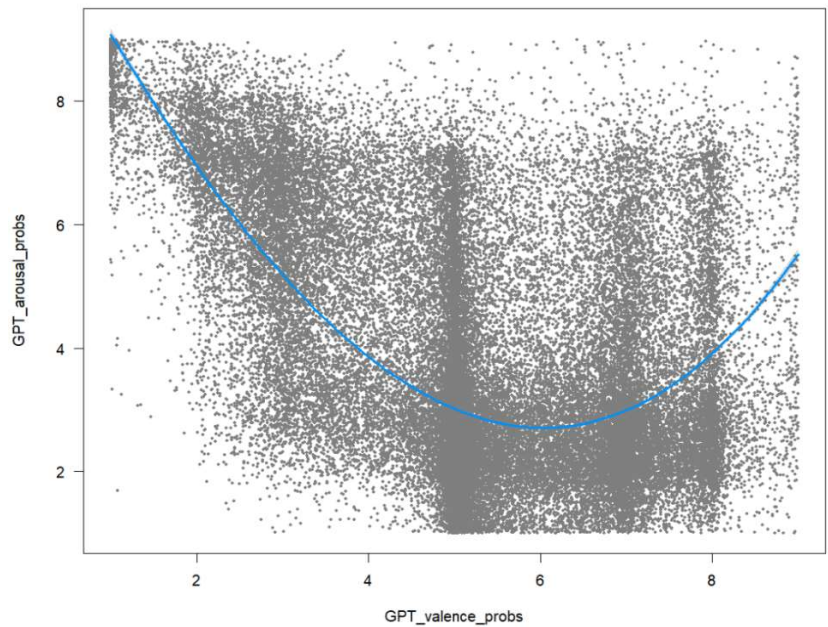

**Figure 4: Correlation between the GTP arousal estimates and the valence estimates for the multiword expressions.**

At the same time, there are exceptions that help us evaluate the quality of the GPT estimates. The 18 expressions with valence estimates between 4 and 6 and arousal estimates above 8.8 were all related to sexuality (sex bomb, sexual pleasure, Latin lover, sexual relationship, sex talk, sexual relation, love affair, …). They were arousing but not necessarily in a positive or negative way, also because there is a gender difference for these words, with men rating such words higher on valence than women (Warriner et al., 2013).

The seven negative expressions with valence estimates below 1.6 and arousal estimates lower than 4 were all infrequent expressions (hare lip, coon bear) and contained outdated slurs (to be found at osf).

The seven positive expressions with valence estimates above 8.999 and arousal estimates below 4 were related to positive experiences that are relaxing rather than arousing (best friend, summer vacation, perfect in every way, vacation time, Merry Christmas and a Happy New Year, summer holiday, Heaven on earth).

All in all, the estimates seem sensible and in line with the high correlations found in Studies 1 and 2. They will certainly allow researchers to search for stimuli in a targeted way.

Figure 5 shows the distributions of the GPT estimates obtained. They resemble the word distributions shown in Figures 2 and 3. The valence estimates are heavily centered on the mean and most expressions are at the low arousal end. Using ranks of the probability-based estimates can tease the values further apart (if that is desired).

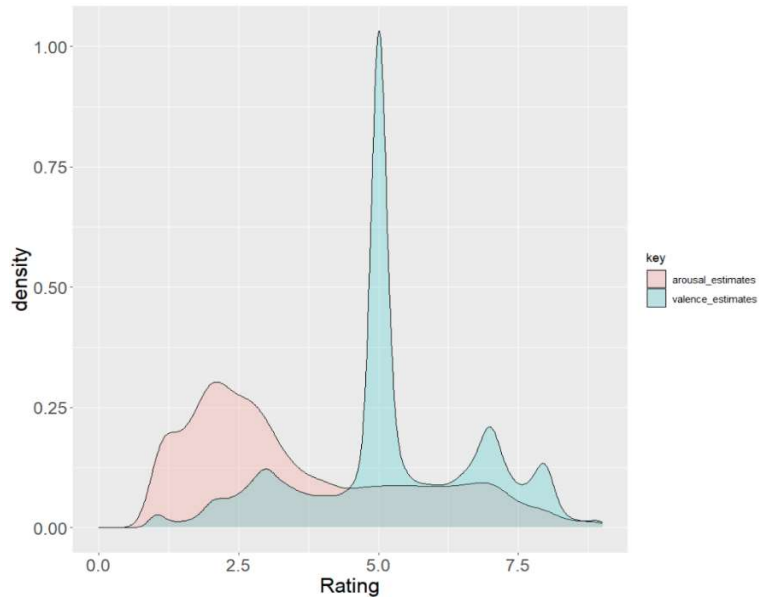

**Figure 5: Distribution of the arousal and valence ChatGPT-4o estimates for the multiword expressions of Muraki et al. (2023).**

## Discussion and availability

The purpose of this article was to see if good estimates of valence and arousal for multi-word expressions can be obtained with large language models. Such estimates cannot be obtained with other artificial intelligence tools because combinations of words often have different meanings than the sum of the individual words and because multi-word expressions can take many forms given that words in them can be inflected, derived or omitted depending on the context in which the expression occurs.

We addressed the problem in a systematic way. First (in Study 1), we investigated whether it was possible to obtain concreteness estimates for multiword expressions that correlated well with the human ratings collected by Muraki et al. (2023). We found that ChatGPT-4o gave good estimates and that adding a few examples of words at each end of the continuum slightly improved the results.[6] Correlation with the human ratings was r = .8. At the same time, the distribution of AI estimates differed from that of the human ratings (Figure 1).

Next (in Study 2), we obtained the same promising results when we tried to estimate the Warriner et al. (2013) word ratings of valence and arousal. The ChatGPT-4o estimates were as good as or better than the estimates obtained with other artificial intelligence tools (Tables 2 and 3).

---

[6] Researchers who disagree with these choices, are of course free to try out other prompts and compare them to the data we obtained. For instance, we kept close to the human ratings and the existing literature by collecting concreteness estimates on a 1-5 scale but valence and arousal estimates on a 1-9 scale. Authors may have reasons to prefer estimates on the same scale.

Finally (in Study 3), we collected valence and arousal ratings for multiword expressions. Although we did not have a large-scale human dataset as a criterion, an evaluation of the extreme words shows that the estimates are as expected. We see no arguments why they would be less good than the concreteness estimates.

Showing that ChatGPT-4o provides good estimates of concreteness, valence, and arousal for words and multiword expression is interesting, because it allows researchers to obtain values for the stimulus materials they are interested in, also materials not covered here.[7] The estimates need not be limited to English, given that ChatGPT-4o is available for many languages. At present, the outcome in these languages is not as good as in English (Martinez et al., 2023), but this is likely to improve in the near future.

Because there is an environment cost if everyone has to run the same analyses over and over again, we make our concreteness, valence, and arousal estimates available in easy to use Excel files. There is a file for the 63,680 multiword expressions, and a file for 126,397 words. The latter was obtained by combining the list of Brysbaert et al. (2019) with the lists of Gao et al. (2023), Scott et al. (2019), and Hollis et al. (2017). This master list includes some faulty entries (mainly from Hollis et al.), but should provide estimates for nearly all words researchers are interested in.

For each variable, the lists contain four columns with:

- The dominant estimate returned by ChatGPT-4o (on a 5-point scale for concreteness[8] and on a 9-point scale for valence and arousal)
- The more precise estimate based on the probabilities of the ratings.
- The relative rank of the stimulus, going from 0 to 1 (i.e., the rank of the stimulus based on the probabilities estimate, divided by the total number of stimuli in the list).
- The rank of the stimulus, going from 1 to 100 (by rounding up the relative ranks), which may be handier for stimulus selection.

All listings are available at https://osf.io/k5a4x/. They can be freely used for research and education, but not for commercial purposes (creative commons license CC BY-NC-SA).

## Declarations

- Funding: This research was supported by the FUN4DATE (PID2022-136684OB-C21/C22) and ENTRUDIT (TED2021-130118B-I00) projects funded by the Spanish Agencia Estatal de Investigacion (AEI) 10.13039/501100011033 and by the OpenAI research access program, which provided access to ChatGPT-4o on a non-commercial basis.
- Conflicts of interest: The authors ran the studies independently and do not expect any financial gain from them.

---

[7] We tested whether the estimates differ as a function of the order of words given to the model. We compared the multiword concreteness ratings for a random list and a list ordered according to the Muraki et al. (2023) ratings going from abstract to concrete. The GPT estimates were the same.

[8] The GPT estimates correlate r = .89 with the word concreteness ratings of Brysbaert et al. (N = 34,246), which compares well to Hollis et al. (r = .83) and Thompson & Lupyan (r = .86).

- Ethics approval: The studies did not involve people and followed the General Ethical Protocol of the Faculty of Psychology and Educational Sciences at Ghent University. Therefore, they need no explicit approval from the Faculty.
- Consent to participate: The studies did not involve new data collection from people.
- Consent for publication: All authors consent.
- Availability of data and materials: All data and materials are available at https://osf.io/k5a4x/.
- Code availability: The R code used for the analyses is available at https://osf.io/k5a4x/.
- Authors' contributions: All authors have contributed to the ideas tested in the paper (and others that did not make the end report). Running the tests was done by the authors from Madrid. Gent was ultimately responsible for statistical analysis and writing.